\documentclass[12pt,twoside]{article}   
\usepackage[super,sort,comma]{natbib}

\usepackage{fancyhdr}
\usepackage{booktabs} 

\newcommand{\captionv}[3]{\begin{center}\parbox{#1cm}{\caption[#2]{{\sf #3}}}
        \end{center}}

\usepackage[section]{placeins}   %

\usepackage{graphicx}
\usepackage{amsmath} 
\usepackage{amssymb}
\usepackage{wrapfig}
\usepackage{multirow}
\makeatletter \renewcommand\@biblabel[1]{$^{#1}$} \makeatother
\newcommand{\cen}[1]{\begin{center} #1 \end{center}}

\usepackage[mathlines]{lineno}

\usepackage{hyperref}
\hypersetup{ colorlinks,
    citecolor=blue,
    filecolor=blue,
    linkcolor=blue,
    urlcolor=blue
}

\usepackage{xcolor}

\definecolor{gray}{rgb}{0.6,0.6,0.6}
\definecolor{red}{rgb}{0.85,0,0}
\definecolor{green}{rgb}{0,0.85,0}
\definecolor{blue}{rgb}{0,0,0.85}
\definecolor{beige}{rgb}{0.92,0.87,0.78}
\usepackage[all]{hypcap}    

\begin{document}

\cen{\sf
{\Large {\bfseries A Neighborhood Attention Transformer Network for
Enhanced 3D Segmentation of the Left Anterior Descending Artery}} \\
\vspace*{6mm}
{\normalsize
Rafi Ibn Sultan\textsuperscript{1},
Chengyin Li\textsuperscript{2},
Yiannos Demetriou\textsuperscript{1},
Ahmed I. Ghanem\textsuperscript{2,3},
Joshua P. Kim\textsuperscript{2},
Justine Cunningham\textsuperscript{2},
Hassan Bagher-Ebadian\textsuperscript{2,4,5,6},
Dongxiao Zhu\textsuperscript{1,7},
and Kundan S. Thind\textsuperscript{2}
}
}

\vspace*{4mm}

\noindent{\small
\textsuperscript{1}Department of Computer Science, Wayne State University, Detroit, MI, 48202, USA \\
\textsuperscript{2}Department of Radiation Oncology, Henry Ford Health, Detroit, MI, 48202, USA \\
\textsuperscript{3}Department of Clinical Oncology, Alexandria University, Alexandria, 21526, Egypt \\
\textsuperscript{4}Department of Radiology, Michigan State University, E.\ Lansing, MI, 48824, USA \\
\textsuperscript{5}College of Osteopathic Medicine, Michigan State University, E.\ Lansing, MI, 48824, USA \\
\textsuperscript{6}Department of Physics, Oakland University, Rochester, MI, 48309, USA \\
\textsuperscript{7}Institute for AI and Data Science, Wayne State University, Detroit, MI, 48202, USA
}

\pagenumbering{roman}
\setcounter{page}{1}
\pagestyle{plain}

\vspace*{3mm}

\noindent{\small
\textbf{Corresponding author:} Rafi Ibn Sultan, Department of Computer Science, Wayne State University,
Detroit, MI, 48202, USA.
E-mail: \href{mailto:rafis@wayne.edu}{rafis@wayne.edu}.
}

\begin{abstract}
\noindent
\textbf{Background:} Accurate segmentation of the Left Anterior Descending (LAD) artery in 3D free-breathing, non-contrast CT is critical for cardiac dose sparing in thoracic radiotherapy. The task is inherently difficult because the LAD is extremely small, exhibits poor soft-tissue contrast, and varies substantially across patients. Even manual contours show limited inter-observer agreement on non-contrast CT, underscoring the ambiguity of the vessel boundaries.

\noindent\textbf{Purpose:} To develop a transformer-based segmentation framework that enhances LAD delineation in low-contrast, imbalanced CT using local-global context modeling and uncertainty-guided optimization.

\noindent\textbf{Methods:} We propose \textbf{NA-UNETR}, a 3D transformer-based segmentation model with Neighborhood Attention (NA) and Dilated NA (DiNA) blocks that jointly capture fine structural detail and long-range context. To address the limited availability of annotated LAD artery data, the model is pretrained on 1,000 CTA volumes representing general coronary anatomy and then fine-tuned using a parameter-efficient adaptation strategy on 20 free-breathing institutional CT scans of the LAD artery. A composite Diceâ€“Focal and Hausdorff loss, dynamically balanced via homoscedastic uncertainty, improves overlap and boundary accuracy. Preprocessing included intensity clipping, contrast enhancement, and artery-centric sampling, and postprocessing refined morphology.

\noindent\textbf{Results:} NA-UNETR achieved comparable performance to SOTA with improved efficiency and boundary properties. It reached 45.64\% Dice, 38.16 mm HD95, and 10.01 mm ASD, improving Dice by 3.10 percentage points over nnU-Net and reducing HD95 by 2.96 mm relative to Swin UNETR, while showing the strongest boundary accuracy among all models and demonstrating improved geometric and centerline stability. On ImageCAS, it achieved 79.49\% Dice, 8.89 mm HD95, and 1.02 mm ASD. Ablations confirmed that residual blocks, variable kernels, and uncertainty-weighted loss each contributed to performance stability and boundary precision.

\noindent\textbf{Conclusions:} NA-UNETR effectively balances local precision and global context, enabling accurate segmentation of thin, low-contrast LAD structures. Its integration of Neighborhood Attention, uncertainty-weighted loss, and LoRA-based fine-tuning demonstrates a robust and computationally efficient framework for substructure-level cardiac segmentation in radiotherapy planning.
\end{abstract}


\newpage     


\newpage

\setlength{\baselineskip}{0.7cm}      

\pagenumbering{arabic}
\setcounter{page}{1}
\pagestyle{fancy}

\section{Introduction}

Radiation-induced heart disease (RIHD) is a growing concern in thoracic oncology, particularly for patients undergoing radiotherapy for lung, esophageal, or breast cancers \cite{nilsson2012distribution, correa2007coronary}. RIHD includes both acute (e.g., pericarditis) and late effects (e.g., coronary artery disease, heart failure, arrhythmias), with risks manifesting earlier than expected, sometimes within 5 years of treatment, and persisting for decades \cite{darby2013risk}. Among cardiac substructures, the Left Anterior Descending (LAD) artery is especially susceptible to radiation injury due to its location along the anterior interventricular groove and its role in supplying the anterior wall and most of the left ventricle \cite{rehman2023anatomy}. Radiation doses as low as 10 Gy have been linked to increased coronary artery calcification and ischemic events \cite{patel2018comparing}, while a mean heart dose greater than 10 Gy or a LAD volume receiving 15 Gy (V\textsubscript{15Gy}) exceeding 10 percent significantly increase the risk of major adverse cardiac events in non-small cell lung cancer patients \cite{atkins2021mean}.

Despite its clinical significance, the LAD is rarely segmented in routine radiotherapy workflows, which often rely on whole-heart doseâ€“volume metrics \cite{song2021dose,van2021importance} such as those recommended by QUANTEC \cite{morris2020cardiac}. These measures treat the heart as a uniform organ, potentially overlooking radiosensitive substructures. Studies indicate that dose to regions like the LAD \cite{nieder2007influence}, left ventricle \cite{van2017validation}, and left atrium \cite{vivekanandan2017impact} may better predict long-term morbidity than global heart dose. However, LAD delineation on standard free-breathing, non-contrast, nonâ€“ECG-gated planning CT remains extremely challenging due to its small caliber, variable course, and limited soft-tissue contrast \cite{wang2025reduced,nicolas2021safety,sakyanun2020effect, zhou2014computerized}. The artery often appears faint with ill-defined boundaries and is further obscured by motion artifacts \cite{van2019development}. Inter-observer studies report manual coronary artery Dice scores ranging from 0.10 to 0.53 on non-contrast CT \cite{duane2017cardiac}, underscoring the substantial disagreement even among expert readers.

Several automatic segmentation strategies have been explored for LAD segmentation. Atlas-based methods are simple to implement but depend on image registration and perform poorly on the LAD, with reported Dice Similarity Coefficients (DSC) of 0.09â€“0.27 \cite{kaderka2019geometric, morris2019cardiac, zhou2017cardiac, van2019development}. Their inability to capture anatomical variability or resolve small, low-contrast structures limits their usefulness. Deep learning approaches generally perform better but still struggle on non-contrast CT: U-Net \cite{ronneberger2015unet} and nnU-Net \cite{isensee2021nnu} achieve only about 0.21 DSC for the LAD \cite{saha2023deep, jin2021robustness}, far lower than the values above 0.85 commonly reported for cardiac chambers and large vessels. This disparity stems from the intrinsic difficulty of LAD delineation on non-contrast CT, where faint boundaries, low contrast, and motion artifacts limit the visual cues available for supervision. These factors, together with the limited availability of expert-labeled data, create a pronounced small-data bottleneck for coronary substructure segmentation. Multi-modal strategies using MRI or contrast-enhanced CT improve visibility \cite{morris2020cardiac, summerfield2024enhancing, li2024novel} but remain impractical for standard radiotherapy and diagnostic workflows.

While this work focuses on the LAD, insights from broader coronary artery segmentation remain relevant since the underlying challenges often overlap. Most pipelines employ convolutional neural networks (CNNs), including U-Net and its variants \cite{kaba2023application,pan2021coronary,song2022automatic}, or enhanced designs with attention and multi-scale fusion \cite{dong2023novel,shen2019coronary}, as well as PSPNet-based \cite{chachadi2025automated} and ensemble approaches \cite{park2023selective}. However, CNNs struggle to capture long-range anatomical dependencies that are essential for maintaining vessel continuity \cite{li2023focalunetr,sultan2025bipvl,mgboh2025fluenceformer}. Transformer-based models address this limitation by modeling global context, allowing relationships between distant structures to be learned jointly. Yet, transformer-based approaches have rarely been explored for coronary artery segmentation and, to our knowledge, have not been evaluated for LAD segmentation in non-contrast, free-breathing planning CT, an especially difficult setting with limited annotated data.

We present \textbf{NA-UNETR}, a 3D transformer-based segmentation framework developed to address the small-data regime inherent to LAD segmentation. Its design integrates Neighborhood Attention (NA) \cite{hassani2023neighborhood} for localized feature extraction within a global transformer backbone, enabling improved representation of thin and low-contrast anatomy. To mitigate class imbalance and improve thin-structure detection, we incorporate artery-focused patch sampling and a composite loss that combines Dice-Focal and boundary-aware Hausdorff losses with homoscedastic uncertainty weighting. Crucially, NA-UNETR employs a two-stage transfer learning strategy: pretraining on a large CCTA dataset followed by parameter-efficient fine-tuning using LoRA on clinically curated planning CTs. This approach directly addresses the scarcity of expert-labeled LAD data and provides a principled solution to the small-data bottleneck. Together, these innovations yield a segmentation framework suited for substructure-level cardiac analysis in thoracic radiotherapy.




\section{Methods}
Our task is formulated as a 3D binary segmentation problem, where the model predicts voxel-wise probabilities of the LAD artery from input CT volumes.

\subsection{Preliminaries: Neighborhood Attention (NA)}
NA is a computationally efficient alternative to global self-attention \cite{hassani2023neighborhood} that restricts token interactions to spatially local neighborhoods while preserving contextual modeling capacity. In 3D medical volumes where the target anatomy may occupy only a minute fraction of voxels and exhibit elongated tubular geometry, unrestricted global attention can mix structurally unrelated regions and dilute weak anatomical signals within dominant background responses. By confining each query to a local $k \times k \times k$ window, NA introduces a spatial inductive bias that promotes geometrically coherent responses among adjacent voxels, improving the delineation of thin, low-contrast vascular structures.

Given an input tensor $\mathbf{X}$, the NA mechanism computes attention over a local $k \times k \times k$ window centered around each query location. For each query $\mathbf{q}_i$ at spatial index $i$, attention is computed only with key-value pairs $\{(\mathbf{k}_j, \mathbf{v}_j) \, | \, j \in \mathcal{N}(i)\}$, where $\mathcal{N}(i)$ denotes the set of neighboring indices. The output is
\begin{equation}
\mathrm{NA}(\mathbf{q}_i) = \sum_{j \in \mathcal{N}(i)} \alpha_{ij} \mathbf{v}_j, \quad 
\alpha_{ij} = \frac{\exp\left( \mathbf{q}_i^\top \mathbf{k}_j / \sqrt{d} \right)}{\sum_{l \in \mathcal{N}(i)} \exp\left( \mathbf{q}_i^\top \mathbf{k}_l / \sqrt{d} \right)},
\label{na_equation}
\end{equation}
\noindent where $\mathbf{q}_i$, $\mathbf{k}_j$, and $\mathbf{v}_j$ are query, key, and value vectors obtained via linear projections of $\mathbf{X}$, and $d$ is the dimensionality of the attention head. In practice, this attention operation is applied independently across multiple heads, where $d$ denotes the dimensionality of each individual head.

\paragraph{Dilated Neighborhood Attention (DiNA).}
While NA effectively models local dependencies, its receptive field increases only linearly with depth. DiNA extends this formulation by sampling dilated neighborhoods, enabling the receptive field to expand progressively across layers while maintaining the underlying locality constraint. This gradual expansion supports integration of longer vessel segments across slices without resorting to fully global interactions, thereby balancing longitudinal context aggregation with local structural fidelity.
The resulting output follows the same form as Eq.~\ref{na_equation}, but with the attention computed over $\mathcal{N}_\delta(i)$ instead of $\mathcal{N}(i)$:
\begin{equation}
\mathrm{DiNA}_{\delta,k}(\mathbf{q}_i) = 
\sum_{j \in \mathcal{N}_\delta(i)} 
\alpha_{ij}^{(\delta)} \mathbf{v}_j, \quad
\alpha_{ij}^{(\delta)} = 
\frac{\exp\left( \mathbf{q}_i^\top \mathbf{k}_j / \sqrt{d} + b(i,j) \right)}
{\sum_{l \in \mathcal{N}_\delta(i)} \exp\left( \mathbf{q}_i^\top \mathbf{k}_l / \sqrt{d} + b(i,l) \right)}.
\end{equation}
\noindent Here, $b(i,j)$ denotes a learnable relative positional bias between query position $i$ and key position $j$, which encodes spatial relationships within the attention window. By alternating NA and DiNA layers, the model preserves locality while exponentially expanding its receptive field without additional computational cost, enabling richer integration of short- and long-range context.

\subsection{NA-UNETR: Architecture}
NA-UNETR (illustrated in \autoref{fig:natunetr}) is a 3D encoderâ€“decoder segmentation framework with a transformer-based architecture incorporating neighborhood attention within a UNETR-style backbone \cite{hatamizadeh2022unetr}.

\subsubsection{Encoder}
\begin{figure}[ht]
\begin{center}
\includegraphics[width=\textwidth]{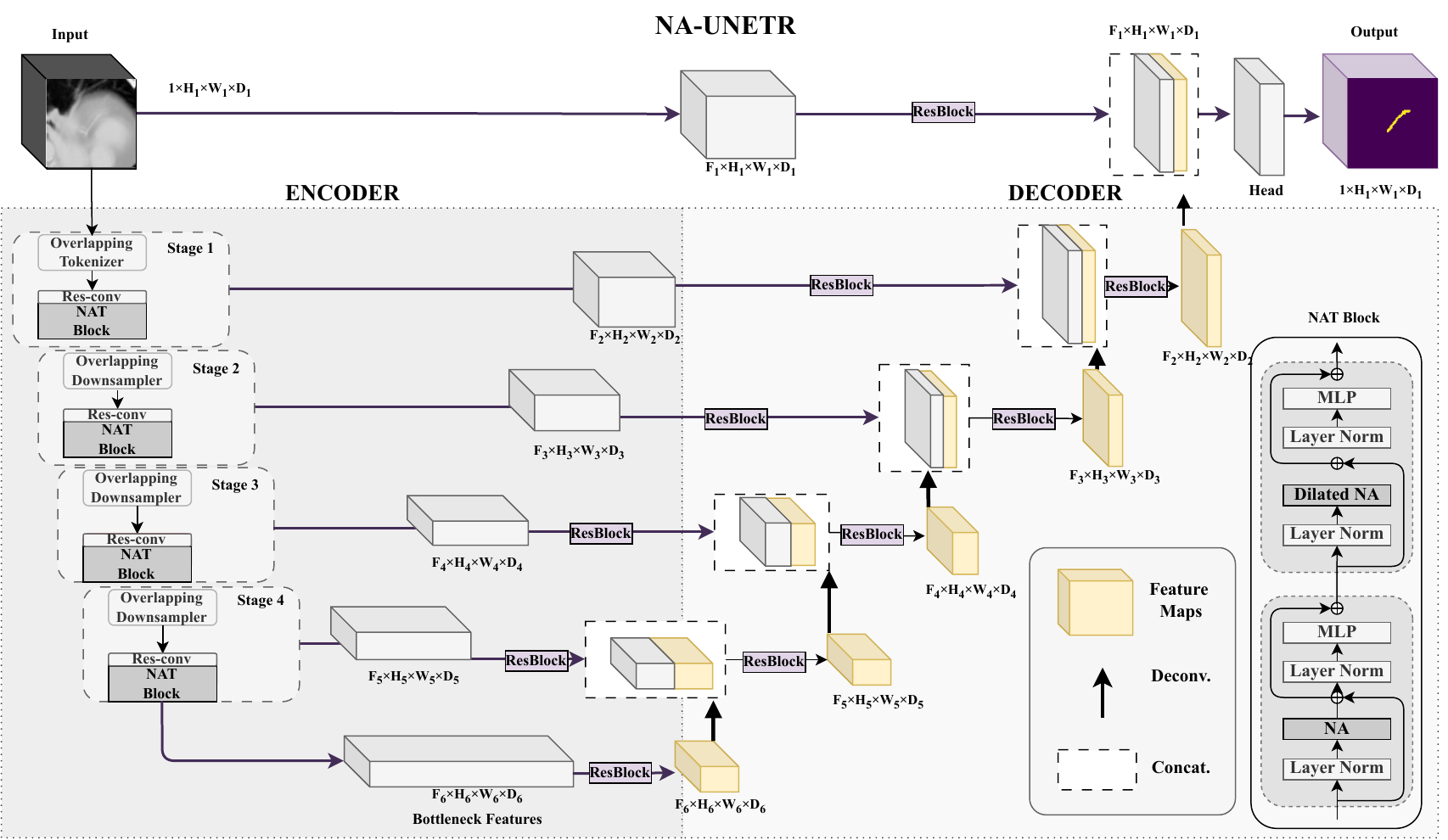}
\captionv{16}{NA-UNETR: Architecture}
{Overall architecture of the proposed NA-UNETR model. The encoder side is divided into four encoder stages, each containing multiple NAT blocks preceded by a residual convolution layer. Skip connections link encoder stages to the corresponding decoder stages, which use residual blocks and upsampling to produce the final segmentation output. Each NAT block can be configured to include DiNA in addition to standard NA, or to use only NA without dilation.
\label{fig:natunetr}}
\end{center}
\end{figure}

As shown in \autoref{fig:natunetr}, NA-UNETR processes a 3D input image $\mathbf{X} \in \mathbb{R}^{1\times H \times W \times D}$, where $(H, W, D)$ are the spatial dimensions and the single input channel indicates a grayscale CT volume. The embedding dimension $d$ in the encoder is set to 48. The visual encoder begins with an overlapping tokenizer consisting of two consecutive $3 \times 3 \times 3$ convolutions with strides of $2 \times 2 \times 2$ and $1 \times 1 \times 1$, which reduce spatial resolution while introducing inductive bias and local connectivity. To enhance locality modeling, depthwise $3 \times 3 \times 3$ convolutions are also used within the tokenizer and feed-forward networks of each transformer block to preserve fine-grained anatomical details.

The encoder is divided into four sequential encoder stages followed by a bottleneck stage (left of \autoref{fig:natunetr}). Each stage contains multiple Neighborhood Attention Transformer (NAT) blocks, which form the core processing units. A residual convolution (Res-Conv) layer is placed before each NAT block to stabilize training and improve gradient flow. Within each NAT block, Neighborhood Attention captures local spatial dependencies within a defined neighborhood, while DiNA can optionally extend this receptive field by introducing a dilation factor. Each NAT block consists of an NA or DiNA layer followed by Layer Normalization (LN), a Multi-Layer Perceptron (MLP), and a second LN. The number of NAT blocks per stage (including the bottleneck stage) is empirically set to (3, 4, 6, 18, 5), and kernel sizes to (7, 7, 7, 3, 3), following \cite{hassani2023neighborhood}. Each stage is preceded by an overlapping downsampler (except the final one), which halves the spatial resolution and doubles the feature channels. The resulting feature maps at each stage are denoted as $(F_2,H_2,W_2,D_2)$, $(F_3,H_3,W_3,D_3)$, $(F_4,H_4,W_4,D_4)$, and $(F_5,H_5,W_5,D_5)$, where $F$ represents the number of channels that increase by $2n$, and $H$, $W$, and $D$ decrease by $2n$ at each successive stage. At each stage, the refined feature map is forwarded to the next stage and retained for decoder skip connections. The final encoder output (bottleneck features) serves as the input to the decoder.

\subsubsection{Decoder}
The decoder in NA-UNETR follows a symmetric U-shaped design (right of \autoref{fig:natunetr}), progressively reconstructing the spatial resolution while integrating semantic information from the encoder through skip connections. Feature maps from each encoder stage are refined by a residual block with two $3 \times 3 \times 3$ convolutions and instance normalization, preserving spatial resolution and channel dimensionality. These refined maps are then upsampled by a deconvolution (transposed convolution) layer that doubles the spatial resolution. Each upsampled feature map is concatenated with its corresponding encoder feature map at the same resolution to restore spatial detail. The concatenated features are further processed by another residual block to fuse contextual and spatial information. After all decoding stages, the resulting feature maps are aggregated through a final convolutional head composed of a $1 \times 1 \times 1$ convolution followed by a sigmoid activation, producing the voxel-wise probability map of the LAD artery. The use of skip connections, deconvolution layers, and concatenation operations ensures that the decoder combines high-level semantic understanding with low-level spatial precision to produce anatomically coherent segmentation outputs.

\subsection{Datasets}

\subsubsection{Institutional Dataset: LAD-SEG}
\label{dataset:lad_seg}
We utilize LAD-SEG, an IRB-approved dataset of free-breathing 3D CT scans from 20 lung cancer patients, each with physician-delineated ground-truth contours of the LAD artery. The scans have a spatial resolution of $(1.17,\,1.17,\,3.0)$~mm and an in-plane size of $512\times512$ voxels, with slice counts ranging from $102$ to $142$. The LAD is represented by small, thin structures within the volume, with an average foreground ratio of $1.7\times 10^{-5}$ and an average of $540$ foreground voxels per case (ranging from $312$ to $1087$ voxels). In terms of intensity characteristics, the foreground Hounsfield Unit (HU) range spans from -669 to 269, with the 1\% and 99\% intensity percentiles lying between -238 and 150. Levene's test for homogeneity of variance, computed across three key attributes: voxel intensity, artery size, and boundary complexity, yielded a statistic of $0.1619$ with a $p$-value of $0.8511$, indicating no statistically significant variance differences among the samples. 

\subsubsection{Public Dataset: ImageCAS}
We leverage the ImageCAS dataset \cite{zeng2023imagecas}, a largeâ€‘scale, publicly available benchmark 
\begin{wrapfigure}{r}{0.5\textwidth}  
\includegraphics[width=0.45\textwidth]{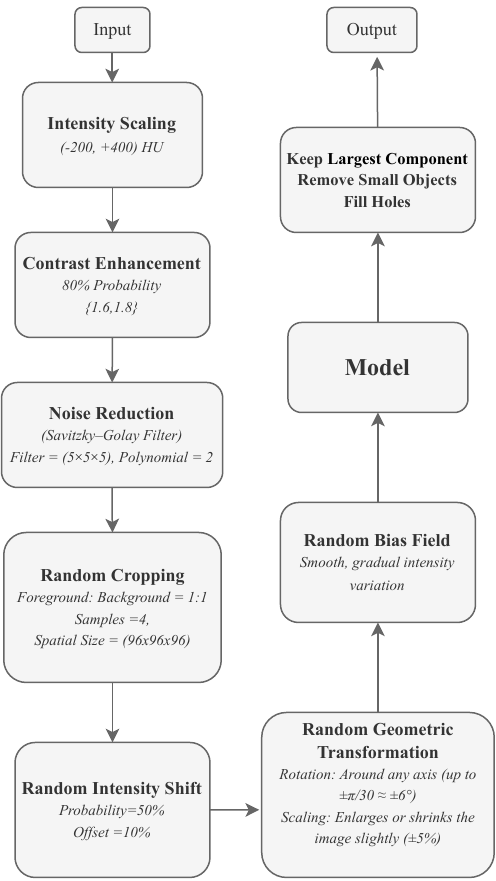}
\vspace{-10pt}
\captionv{8}{Data processing techniques}
{Overview of preprocessing, postprocessing, and augmentation used to address severe class imbalance and enhance vessel-to-background contrast in non-contrast CT scans. \label{fig:processing}}
\vspace{-20mm}
\end{wrapfigure}
specifically designed for coronary artery segmentation from computed tomography angiography (CTA) images. ImageCAS consists of 1,000 highâ€‘resolution 3D CTA volumes acquired using Siemens 128â€‘slice dualâ€‘source scanners at Guangdong Provincial Peopleâ€™s Hospital between 2012 and 2018. Each scan (512Ã—512Ã—206â€“275 voxels, 0.29â€“0.43 mm inâ€‘plane resolution, 0.25â€“0.45 mm slice spacing) includes meticulous voxelâ€‘wise annotations of both left and right coronary arteries by two expert radiologists, with crossâ€‘validation and adjudication by a third in case of disagreement. Only binary coronary artery masks are provided in the dataset, without further subclass labels or vesselâ€‘type categories.

\subsection{Data Preprocessing}
\label{sec:processing}
We apply a preprocessing pipeline tailored to the LAD-SEG dataset to address class imbalance and low vessel-to-background contrast, as illustrated in \autoref{fig:processing}.

\subsubsection{Addressing Data Imbalance}
To alleviate severe class imbalance, training patches of size $(96,96,96)$ were extracted based on the typical spatial extent and morphology of coronary arteries. For each training iteration, an equal number of patches were randomly cropped around positive (LAD) and negative (background) regions in a 1:1 ratio, ensuring balanced exposure of both classes during optimization. Four random patches were generated per image in each iteration, promoting robust learning of small vessel structures while preventing bias toward background regions.

\begin{wrapfigure}{r}{0.5\textwidth}  
\includegraphics[width=0.5\textwidth]{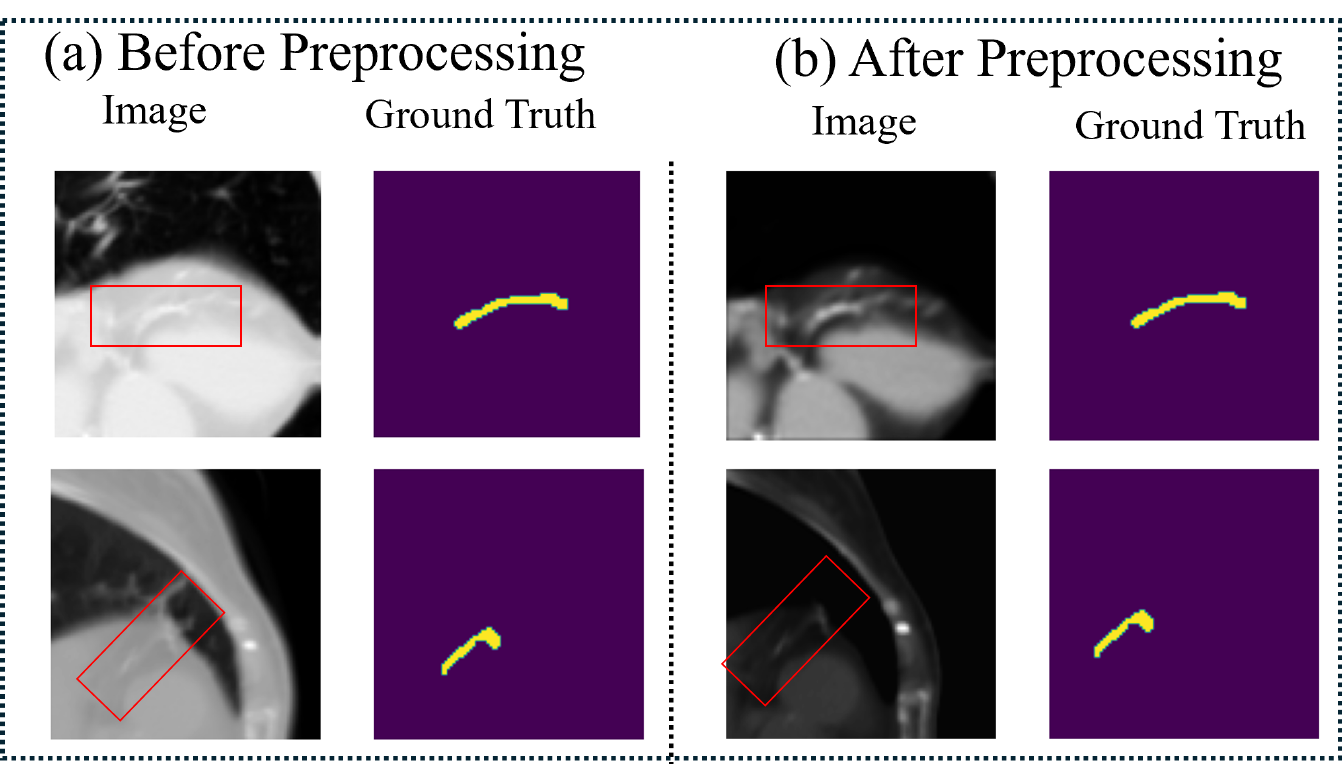}
\vspace{-10pt}
\captionv{8}{NAT vs DiNAT Blocks}
{LAD artery segmentation without preprocessing (left) and with preprocessing (right). Each shows a low-contrast CT patch and its predicted mask overlaid on the ground truth (yellow). Red boxes highlight regions where preprocessing improves vessel edge clarity and continuity. \label{fig:preprocessing}}
\end{wrapfigure}

\subsubsection{Addressing Low Contrast}
LAD intensity statistics in Section~\ref{dataset:lad_seg} show that vessel voxels lie primarily within a narrow soft-tissue range, with most values between the 1st and 99th percentiles of \(-238\) and \(150\)~HU. Accordingly, we clipped CT intensities to \([-200, 400]\)~HU to standardize contrast and suppress unrelated high-density regions. Following clipping, all intensity values were linearly rescaled to the range $[0, 1]$. Local contrast was randomly enhanced using a gamma adjustment factor between 1.6 and 1.8 (applied with probability 0.8), while random intensity shifts promoted invariance to scanner-specific baseline differences. A Savitzkyâ€“Golay filter with a window length of 5 and polynomial order of 2 was applied for smoothing, followed by random global intensity shifts of up to $\pm 0.10$ with a probability of 0.5. The qualitative impact of these preprocessing steps is illustrated in \autoref{fig:preprocessing}, where the preprocessed images show improved vessel visibility and continuity in low-contrast regions compared to the non-preprocessed inputs.

\subsubsection{Geometric and Structural Augmentations}
Training images and labels were randomly rotated (up to $\pi/30$ radians, $\pm 6^{\circ}$) and scaled (up to $\pm 5\%$) while maintaining the target spatial size. These small perturbations simulated anatomical and positioning variability without distorting vessel morphology. Random Gaussian sharpening with variable kernel widths enhanced local edge details, while low-magnitude bias fields introduced smooth, gradual low-frequency intensity variations to simulate scanner-induced inhomogeneities and improve generalization to clinical data.

\subsection{Loss Function}

We introduce a loss formulation that integrates a segmentation loss with a boundary-aware loss to better refine the contours of thin structures. The two components are balanced using homoscedastic uncertainty weighting \cite{kendall2018multi}, allowing their relative contributions to adjust dynamically during training.

\subsubsection{Segmentation Loss: Dice-Focal Loss}
We adopt Dice-Focal loss as the primary segmentation loss, which combines the regionâ€‘overlap sensitivity of Dice Loss \cite{milletari2016v} with the hardâ€‘example focusing of Focal Loss \cite{lin2017focal}, making it more suitable for highly imbalanced anatomical structures. Given the predicted probability map $\hat{\mathbf{Y}}$ and the ground truth label map $\mathbf{Y}$, the Dice component is formulated as:
\begin{equation}
\mathcal{L}_{\text{Dice}} = 1 - \frac{2 \sum_{n} y_n \hat{y}_n + \epsilon}{\sum_{n} y_n + \sum_{n} \hat{y}_n + \epsilon},
\end{equation}
where $y_n$ and $\hat{y}_n$ denote the ground truth and prediction at voxel $n$, and $\epsilon$ is a small constant for numerical stability.

The focal component modulates the standard cross-entropy term by down-weighting well-classified examples and focusing on harder ones:
\begin{equation}
\mathcal{L}_{\text{Focal}} = - \frac{1}{N} \sum_{n} \alpha (1 - \hat{y}_n)^{\gamma} y_n \log(\hat{y}_n) + (1-\alpha) \hat{y}_n^{\gamma} (1 - y_n)\log(1-\hat{y}_n),
\end{equation}
where $\alpha$ balances positive and negative samples and $\gamma$ controls the focusing strength.

\noindent The combined Dice-Focal Loss is then defined as:
\begin{equation}
\mathcal{L}_{\text{Dice-Focal}} = \lambda_1 \mathcal{L}_{\text{Dice}} + \lambda_2 \mathcal{L}_{\text{Focal}},
\end{equation}
with $\lambda_1$ and $\lambda_2$ controlling the contribution of each component. This loss guides the network toward high overlap with ground truth while emphasizing difficult voxels along the vessel.

\subsubsection{Boundary Loss: Hausdorff Loss}
To improve boundary alignment alongside segmentation accuracy, we adopt Hausdorff Loss \cite{karimi2019reducing} based on the Hausdorff distance, which explicitly penalizes large deviations between predicted and true boundaries, improving boundary precision. 

Let $\partial \hat{\mathbf{Y}}$ and $\partial \mathbf{Y}$ denote the sets of boundary points extracted from the predicted segmentation $\hat{\mathbf{Y}}$ and the ground truth $\mathbf{Y}$. The symmetric Hausdorff distance between them is defined as:
\begin{equation}
d_H(\partial \hat{\mathbf{Y}}, \partial \mathbf{Y}) = \max \left\{ \sup_{x \in \partial \hat{\mathbf{Y}}} \inf_{y \in \partial \mathbf{Y}} d(x,y), \ \sup_{y \in \partial \mathbf{Y}} \inf_{x \in \partial \hat{\mathbf{Y}}} d(x,y) \right\},
\end{equation}
where $d(x,y)$ is the Euclidean distance between points $x$ and $y$. In practice, we use a differentiable approximation:
\begin{equation}
\mathcal{L}_{\text{Hausdorff}} = \frac{1}{|\partial \hat{\mathbf{Y}}|}\sum_{x \in \partial \hat{\mathbf{Y}}} \min_{y \in \partial \mathbf{Y}} d(x,y)^2 + \frac{1}{|\partial \mathbf{Y}|}\sum_{y \in \partial \mathbf{Y}} \min_{x \in \partial \hat{\mathbf{Y}}} d(x,y)^2.
\end{equation}

\subsubsection{Loss Balancing Using Homoscedastic Uncertainty}

During training, the contributions of the segmentation loss and the boundary loss are dynamically balanced using homoscedastic uncertainty weighting \cite{kendall2018multi}. This approach enables the model to learn the relative importance of each task, eliminating the need for difficult and task-specific manual weight tuning. Two learnable logâ€‘variance parameters, $\log \sigma_1$ and $\log \sigma_2$, are introduced for the Dice-Focal Loss and the Hausdorff Loss, respectively; these are exponentiated and clamped to obtain the variance terms $\sigma_1^2$ and $\sigma_2^2$. A zero-mean Gaussian noise term $\epsilon \sim \mathcal{N}(0,\sigma_n^2)$ with small variance was added to the Hausdorff Loss to form $\tilde{\mathcal{L}}_{\text{Hausdorff}}$, slowing convergence and encouraging exploration of both global and local boundary features, which in turn prevented overfitting to early boundary estimates and promoted smoother, more consistent boundary learning. The total loss is then computed as  
\begin{equation}
\mathcal{L}_{\text{total}} = \frac{1}{2\sigma_1^2}\,\mathcal{L}_{\text{Dice-Focal}} + \frac{1}{2\sigma_2^2}\,\tilde{\mathcal{L}}_{\text{Hausdorff}} + \log\sigma_1 + \log\sigma_2,
\end{equation}
where $\mathcal{L}_{\text{Dice-Focal}}$ is the combined Dice and focal Loss, $\tilde{\mathcal{L}}_{\text{Hausdorff}}$ is the noiseâ€‘perturbed Hausdorff Loss, $\sigma_1^2$ and $\sigma_2^2$ control the relative weighting of the two losses, and the logarithmic terms act as regularizers.


\subsection{Data Postprocessing}
\label{sec:postprocessing}
To refine the predicted LAD masks and reduce spurious false positives, we applied a three-step postprocessing pipeline. First, only the largest connected component corresponding to the foreground was retained, ensuring that isolated noisy predictions were discarded. Next, connected components smaller than $64$ voxels were removed to suppress small, irrelevant structures. Finally, residual holes within the retained LAD region were filled to produce contiguous, anatomically plausible segmentations. 


\section{Implementation Details}

\subsection{Training Strategy: Pretraining and Fine-tuning}

We first pretrained all models on the ImageCAS dataset \cite{zeng2023imagecas}, which consists of 1,000 3D CTA volumes of coronary arteries. This pretraining provides generalized encoder representations of coronary artery structures that transfer effectively to smaller downstream datasets, such as LAD artery segmentation. Let $\mathcal{D}_{\text{pre}} = \{(\mathbf{X}^{(i)}, \mathbf{Y}^{(i)})\}_{i=1}^{1000}$ denote this dataset, where $\mathbf{X}^{(i)} \in \mathbb{R}^{C \times H \times W \times D}$ are the input volumes and $\mathbf{Y}^{(i)}$ are their corresponding ground-truth segmentations. The parameters of the pretrained encoder and decoder are denoted by $\theta_{\text{enc}}^{*}$ and $\theta_{\text{dec}}^{*}$, respectively.

For fine-tuning on our in-house LAD-Seg dataset $\mathcal{D}_{\text{LAD}}$, all benchmark models and our NA-UNETR were initialized with the pretrained weights ($\theta_{\text{enc}}^{*}$ and $\theta_{\text{dec}}^{*}$). To efficiently adapt these models, we employed Low-Rank Adaptation (LoRA) \cite{hu2022lora}, a parameter-efficient fine-tuning method that adapts large models without retraining all their weights. The attention layers within the encoder were frozen, and we fine-tuned only the decoder parameters ($\theta_{\text{dec}}$) and the inserted LoRA adapters. Specifically, the MLP layers in each encoder block were replaced with LoRA modules of rank $r=8$, defined as $W \gets W + AB$, where $W$ is the original pretrained MLP weight matrix and $A \in \mathbb{R}^{d \times r}$ and $B \in \mathbb{R}^{r \times d}$ are the trainable low-rank factors. During this fine-tuning stage, only $\theta_{\text{dec}}$, $A$, and $B$ were updated, while all other pretrained weights remained fixed.

\subsection{Experiment Setup}  
We designed two experimental approaches based on the two datasets.  
\begin{wrapfigure}{r}{0.5\textwidth}  
\includegraphics[width=0.5\textwidth]{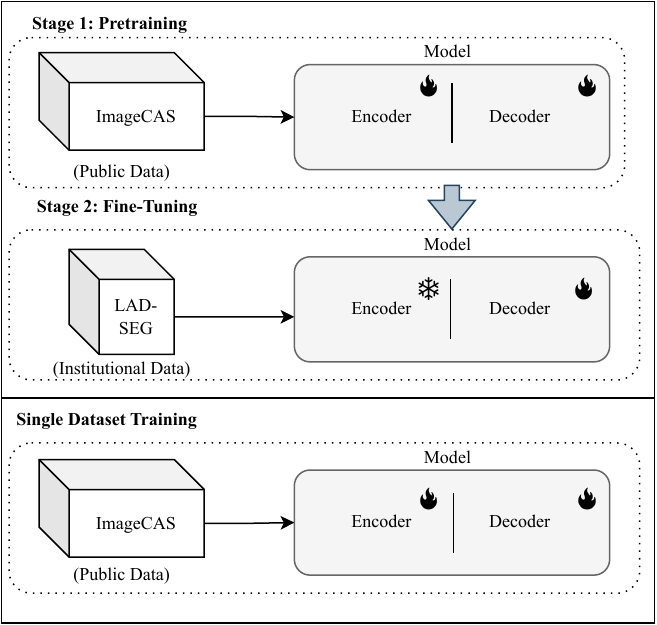}
\vspace{-10pt}
\captionv{8}{Two approaches}
{Overview of the two experimental approaches: the top section shows pretraining on ImageCAS followed by fine-tuning on LAD-SEG, while the bottom section shows direct training on ImageCAS. \label{fig:approaches}}
\vspace{-5mm}
\end{wrapfigure}
The common setup for both is as follows:

All models in all experimental settings were trained using our proposed loss function strategy, with a balancing factor $\alpha = 0.8$ and focusing parameter $\gamma = 2$ for the focal loss component, where class weights were set to 0.1 for the background and 0.9 for the foreground to counter the class imbalance.  The weighting coefficients $\lambda_1$ and $\lambda_2$ were set to $\lambda_1 = 1$ and $\lambda_2 = 1$ in all experiments based on empirical evaluation, as alternative settings (0.5/1, 1/0.5, and 2/1) did not improve DSC or HD95. Class imbalance was handled through the focal loss parameters and class weighting described above.
Optimized with the AdamW optimizer (learning rate $10^{-4}$, weight decay $10^{-5}$). All experiments were implemented in PyTorch 2.5.1 with Python 3.9.21 and executed on a server equipped with eight NVIDIA A100 GPUs (40 GB memory each), with each model trained on a single GPU.  
The specific configurations for each experimental approach (\autoref{fig:approaches}) are detailed below. 
\subsubsection{Pretraining \& Fine-Tuning Approach}  
In this approach, each model was first pretrained on a large public dataset to learn generalized feature representations and then fine-tuned on the smaller in-house dataset to tackle the limited availability of annotated LAD scans for supervised training.

\noindent \textbf{ImageCAS.}  
In the pretraining stage, models were trained on the full ImageCAS dataset, comprising 1,000 images, with 5\% reserved for validation. Training was conducted for 100 epochs, with validation performed every 10 epochs to monitor performance and store the best-performing weights. Preprocessing included clipping voxel intensities to the range $[-200, 500]$ Hounsfield units, followed by linear scaling to $[0, 1]$. To address class imbalance, a 1:1 random cropping strategy was applied between foreground (vessel) and background regions. Additional augmentations included random intensity shifts, random affine transformations, and other standard preprocessing operations. Each benchmark model was trained under this configuration, and the resulting pretrained weights will be made publicly available to facilitate further research.  

\noindent \textbf{LAD-SEG.}  
In the fine-tuning stage, we applied a 5-fold split of the LAD-SEG dataset, with each fold serving as an independent validation set. The dataset was randomly partitioned into five approximately equal folds using a fixed random seed for reproducibility. In each iteration, four folds were used for training and one for validation, with the process repeated until every fold had served as the validation set exactly once. Model predictions were evaluated against physician-delineated ground-truth LAD contours, and final performance was reported as the average across all folds.  Each model (including all baselines and NA-UNETR) was trained for 200 epochs under the same training protocol, using identical LAD-SEG preprocessing and postprocessing procedures described in \autoref{sec:processing} and \autoref{sec:postprocessing}, the same data splits, sampling strategy, loss formulation, and optimization settings. 

\subsubsection{Single-Dataset Direct Training} 
In this approach, both training and evaluation were performed on the public dataset to assess model performance on a well-established benchmark.

\noindent \textbf{ImageCAS.}  
In the single-dataset training setting, models were trained and evaluated exclusively on the ImageCAS dataset to enable direct comparison with benchmark methods on this public dataset. Following the official protocol of \cite{zeng2023imagecas}, we performed 4-fold splits, training for 30 epochs each split. Preprocessing steps were identical to those used in the previous approach.

\subsection{Benchmark Models}
A diverse set of CNN-based and Transformer-based segmentation models that represent state-of-the-art approaches in medical image analysis were selected as benchmark models. Each model followed its official implementation, with dataset-specific tweaks applied (consistent with those used in our proposed model) for fair evaluation. For Transformer-based segmentation backbones (e.g., UNETR), we applied LoRA to the MLP layers within each Transformer block for parameter-efficient fine-tuning. In contrast, for CNN-based models (e.g., U-Net, U-Net++), which lack MLP components, LoRA was not used; instead, we fine-tuned only the decoder while keeping the encoder frozen.

\paragraph{CNN-based Models}
\begin{itemize}
    \item \textbf{U-Net}~\cite{ronneberger2015unet}: Classic encoderâ€“decoder with skip connections, preserving spatial details for biomedical segmentation.
    \item \textbf{UNet++}~\cite{zhou2018unet++}: U-Net variant with nested dense skip connections for improved multi-scale feature fusion.
    \item \textbf{nnU-Net}~\cite{isensee2021nnu}: Self-configuring framework that automatically adapts preprocessing, architecture, and training to datasets.
    \item \textbf{MedNeXt}~\cite{roy2023mednext}: ConvNeXt-inspired CNN using large kernels and residual connections for efficient long-range context capture.
\end{itemize}

\paragraph{Transformer-based Models}
\begin{itemize}
    \item \textbf{UNETR}~\cite{hatamizadeh2022unetr}: Transformer-based encoderâ€“decoder with a ViT encoder and multi-scale feature aggregation in the decoder.
    \item \textbf{Swin UNETR}~\cite{hatamizadeh2021swin}: Swin Transformer backbone using shifted-window self-attention for locality and efficiency.
    \item \textbf{Swin UNETR-V2}~\cite{he2023swinunetr}: Refined Swin UNETR with architectural and training improvements for stable representation learning.
    \item \textbf{nnFormer}~\cite{zhou2021nnformer}: Hybrid CNNâ€“Transformer for 3D segmentation, balancing local convolutions with global attention.
\end{itemize}


\subsection{Evaluation Metrics}
We assess model performance using four primary metrics: Dice Similarity Coefficient (DSC, \%), centerline Dice (clDice, \%), 95th-percentile Hausdorff Distance (HD95, mm), and Average Surface Distance (ASD, mm).

\noindent The Dice Similarity Coefficient (DSC) measures the volumetric overlap between the predicted segmentation map \(P\) and the ground truth mask \(G\), defined as:
\begin{equation}
\text{DSC} = \frac{2 \sum_{j} P_j G_j}{\sum_{j} P_j^2 + \sum_{j} G_j^2} \times 100,
\end{equation}
where \(P_j\) and \(G_j\) denote the predicted probability and ground truth value for voxel \(j\). Higher DSC indicates better overlap.

\noindent To complement DSC for thin tubular structures, we also report centerline Dice (clDice), which evaluates topological agreement by comparing the skeletons of the prediction and ground truth. Let \(S_P = \mathrm{skel}(P)\) and \(S_G = \mathrm{skel}(G)\) denote their centerline skeletons. clDice is defined as:
\begin{equation}
\text{clDice} =
\frac{2T_{\mathrm{prec}}T_{\mathrm{sens}}}
{T_{\mathrm{prec}}+T_{\mathrm{sens}}}\times100,
\quad
T_{\mathrm{prec}}=\frac{|S_P\cap G|}{|S_P|},
\quad
T_{\mathrm{sens}}=\frac{|S_G\cap P|}{|S_G|}.
\end{equation}
Higher clDice indicates better preservation of the vessel trajectory.

\noindent The 95th-percentile Hausdorff distance (HD95) evaluates the geometric alignment between the predicted segmentation boundary and the ground truth, defined as:
\begin{equation}
\text{HD95} = \operatorname{quantile}_{95\%}\! \left(\max_{x \in \partial \hat{p}} \min_{y \in \partial G} \|x - y\|\right),
\end{equation}
where \(\partial \hat{p}\) and \(\partial G\) denote predicted and ground-truth boundaries. Lower HD95 indicates fewer large boundary deviations.

\noindent The Average Surface Distance (ASD) quantifies mean symmetric boundary distance:
\begin{equation}
\text{ASD} = \frac{1}{|\partial \hat{p}| + |\partial G|}
\left( 
\sum_{x \in \partial \hat{p}} \min_{y \in \partial G} \|x-y\| +
\sum_{y \in \partial G} \min_{x \in \partial \hat{p}} \|y-x\|
\right),
\end{equation}
with lower values indicating closer average alignment between predicted and true boundaries.

\noindent To assess the statistical significance of performance differences, we conducted the non-parametric Mann--Whitney U test on the per-case metric values across all methods. The test was applied separately to the results from both datasets (our in-house LAD-SEG dataset and the public ImageCAS dataset) for DSC, HD95, and ASD. Reported $p$-values indicate whether the improvements achieved by NA-UNETR over competing baselines are statistically significant ($p < 0.05$).

\begin{figure}[ht]
\begin{center}
\includegraphics[width=0.9\textwidth]{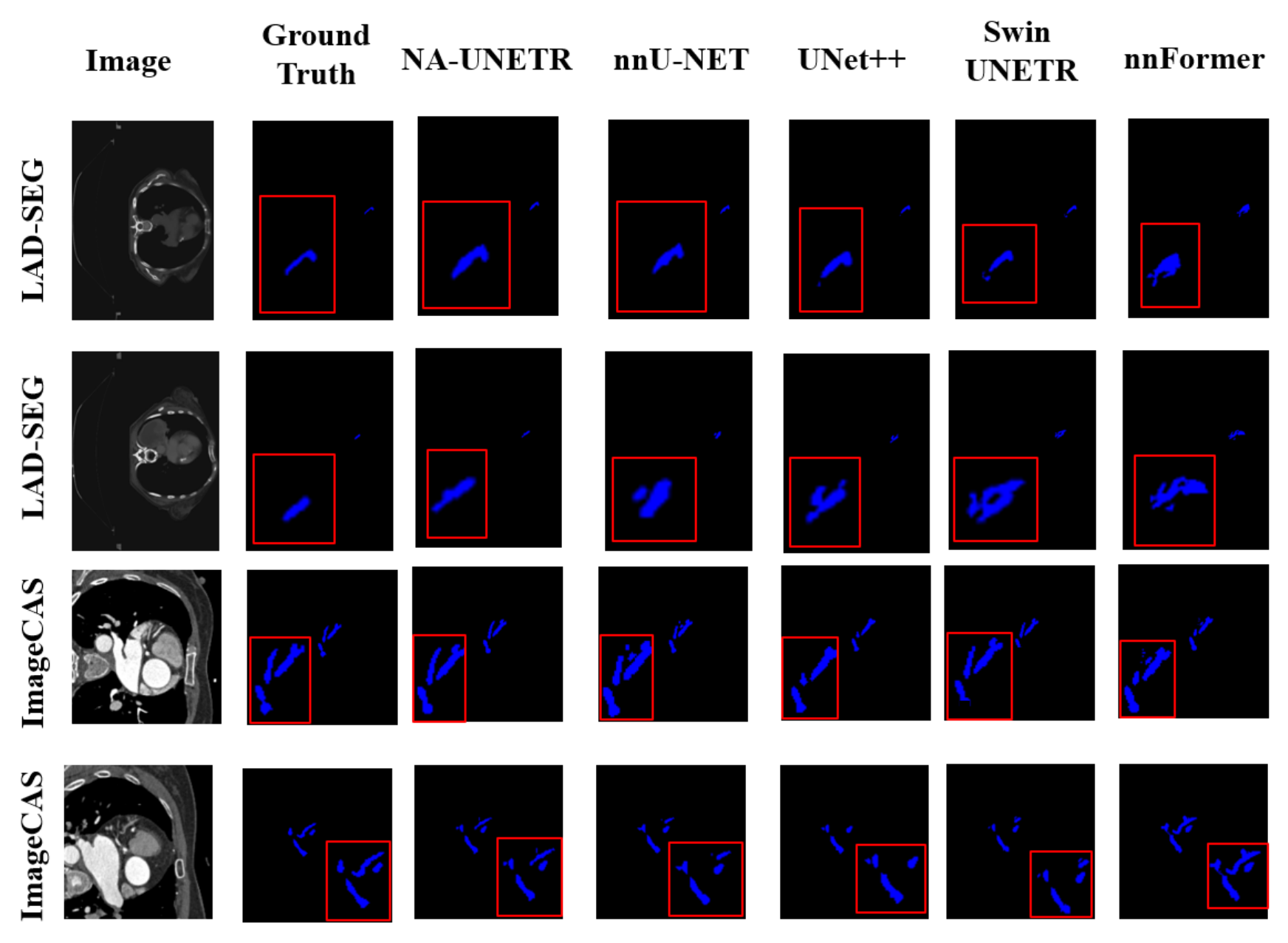}
\captionv{16}{NA-UNETR: Architecture}
{Representative qualitative results on randomly selected slices from the LAD-SEG and ImageCAS validation datasets. Each row corresponds to a randomly chosen slice from a randomly selected case. The first column shows the CT image, followed by the ground truth annotation, and predictions obtained using the proposed NA-UNETR (ours) and other state-of-the-art (SOTA) segmentation models. To facilitate visual comparison of the vascular structures, the red bounding boxes indicate manually selected regions containing the target vessel and are shown as zoomed-in views of the predicted masks.}
\label{fig:visual}
\end{center}
\end{figure}

\section{Results}
\subsection{Qualitative Results}
\begin{wrapfigure}{r}{0.5\textwidth}
\vspace{-10pt}
\centering
\includegraphics[width=\linewidth]{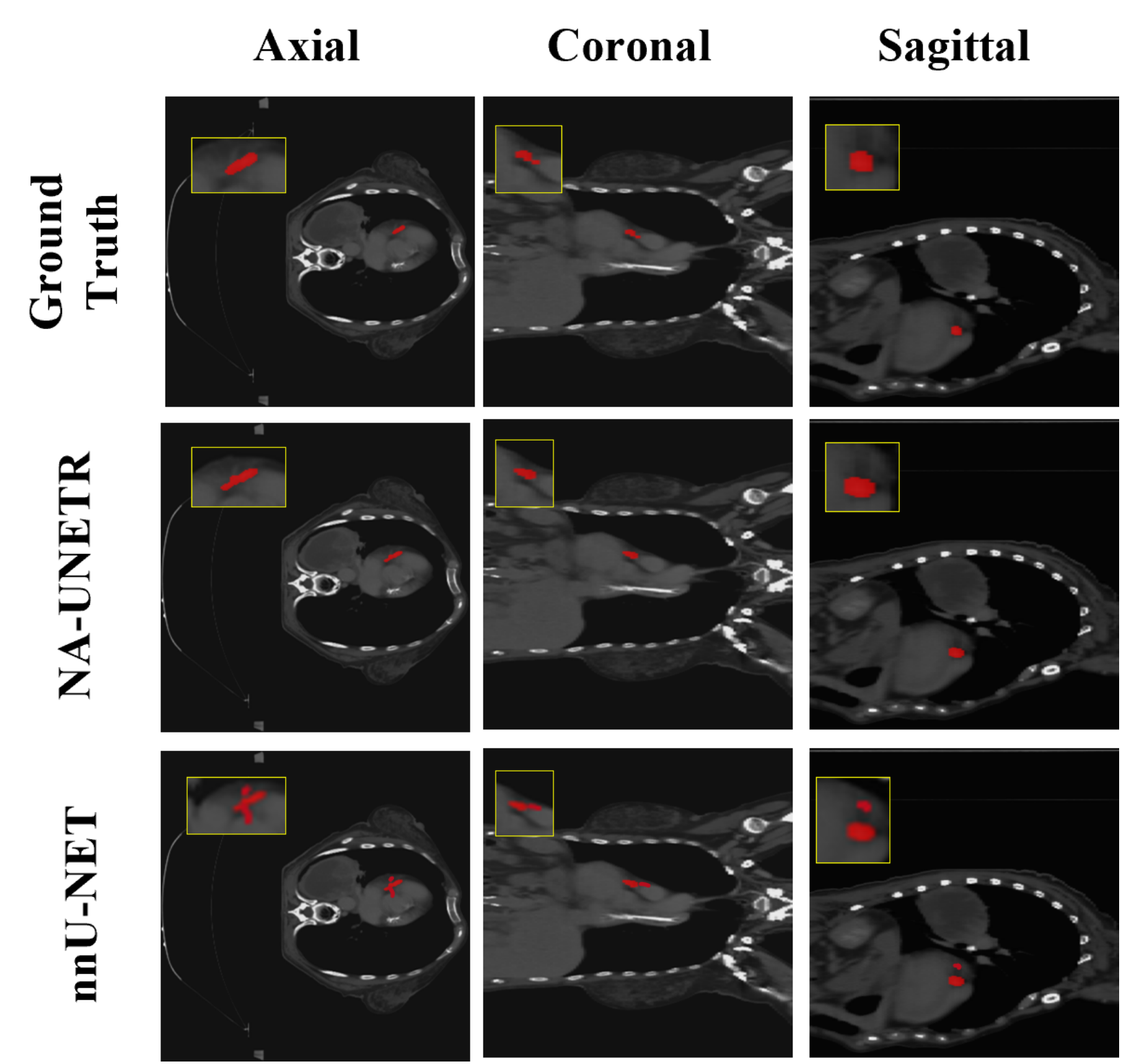}
\vspace{-20pt}
\caption{Representative axial, coronal, and sagittal views with zoomed-in ROI insets (yellow boxes) highlighting the segmented LAD region across Ground Truth, NA-UNETR, and nnU-Net predictions.}
\label{fig:multiviews}
\end{wrapfigure}

Qualitative comparisons on randomly selected slices from both LAD-SEG and ImageCAS datasets (\autoref{fig:visual}) demonstrate that NA-UNETR provides clearer and more accurate delineations of the coronary arteries compared with benchmark models. 
Each prediction is accompanied by a zoomed-in view (red bounding box) to highlight subtle differences in boundary quality. As shown, NA-UNETR consistently produces vessel contours that align more closely with the ground truth, whereas competing SOTA methods often exhibit fragmented boundaries, spurious predictions, or missed segments. These improvements are particularly pronounced on the LAD-SEG dataset, which is more challenging due to its low contrast and thin vessel boundaries. In such cases, NA-UNETR preserves structural continuity and reduces noise, offering more reliable vessel localization. To further illustrate the three-dimensional structural consistency of NA-UNETR and its comparison with nnU-Net, representative axial, coronal, and sagittal views are presented in \autoref{fig:multiviews}.

\subsection{Quantitative Results}

\begin{table}[htbp]
\begin{center}
\captionv{15}{tab:main_results}{Quantitative results of NA-UNETR against different SOTA models on our institutional data LAD-SEG (both CNN-based and Transformer-based). Each entry reports the mean Â± standard deviation. NA-UNETR (no-dil.) is NA-UNETR with NA blocks instead of DiNA blocks. The best results are bolded.
\label{tab:main_results}
\vspace*{2ex}
}
\begin{tabular}{l|c|c|c|c}
\hline
\textbf{Method} & \textbf{DSC (\%)} $\uparrow$ & \textbf{clDice (\%)} $\uparrow$ & \textbf{HD95 (mm)} $\downarrow$ & \textbf{ASD (mm)} $\downarrow$ \\
\hline
U-Net~\cite{ronneberger2015unet}       & 35.30$\pm$5.91 & 37.20$\pm$11.31 & 43.80$\pm$7.55 & 12.80$\pm$3.27 \\
UNet++~\cite{zhou2018unet++}           & 35.57$\pm$8.24 & 42.40$\pm$10.43 & 40.95$\pm$6.05 & 12.09$\pm$1.91 \\
nnU-Net~\cite{isensee2021nnu}          & 42.54$\pm$2.90 & 42.33$\pm$10.97 & 39.68$\pm$5.92 & 10.37$\pm$1.56 \\
MedNeXt~\cite{roy2023mednext}          & 38.79$\pm$6.44 & 40.91$\pm$10.49 & 41.68$\pm$4.80 & 11.78$\pm$2.56 \\
\hline
UNETR~\cite{hatamizadeh2022unetr}      & 42.13$\pm$3.55 & 42.08$\pm$8.70 & 40.65$\pm$6.96 & 10.44$\pm$1.26 \\
Swin UNETR~\cite{hatamizadeh2021swin}  & 44.78$\pm$6.08 & 43.76$\pm$9.33 & 41.12$\pm$5.98 & 10.79$\pm$2.45 \\
SwinUNETR-V2~\cite{he2023swinunetr}    & 44.11$\pm$4.54 & 43.50$\pm$9.79 & 40.29$\pm$7.28 & 10.18$\pm$2.10 \\
nnFormer~\cite{zhou2021nnformer}       & 42.04$\pm$5.34 & 41.88$\pm$9.71 & 40.43$\pm$7.11 & 11.05$\pm$3.02 \\
\hline
NA-UNETR (no-dil.)               & 44.09$\pm$3.89 & 43.45$\pm$9.06 & 40.10$\pm$9.30 & \textbf{9.55$\pm$3.24} \\
NA-UNETR (ours)                        & \textbf{45.64$\pm$4.86} & \textbf{44.39$\pm$8.38} & \textbf{38.16$\pm$4.37} & 10.01$\pm$1.39 \\
\hline
\hline
\end{tabular}
\end{center}
\end{table}

\begin{table}[htbp]
\begin{center}
\captionv{15}{tab:main_results2}{Quantitative results of NA-UNETR against different SOTA models on the public data ImageCAS dataset (both CNN-based and Transformer-based). NA-UNETR (no-dil.) is NA-UNETR with NA blocks instead of DiNA blocks.  Each entry reports the mean Â± standard deviation. The best results are bolded.
\label{tab:main_results2}
\vspace*{2ex}
}
\begin{tabular}{l|c|c|c|c}
\hline
\textbf{Method} & \textbf{DSC (\%)} $\uparrow$ & \textbf{clDice (\%)} $\uparrow$ & \textbf{HD95 (mm)} $\downarrow$ & \textbf{ASD (mm)} $\downarrow$ \\
\hline
U-Net~\cite{ronneberger2015unet}       & 73.47$\pm$0.32 & 82.07$\pm$0.49 & 10.69$\pm$0.78 & 1.39$\pm$0.06 \\
UNet++~\cite{zhou2018unet++}           & 78.57$\pm$0.51 & 84.71$\pm$0.69 & 9.46$\pm$0.84  & 1.15$\pm$0.07 \\
nnU-Net~\cite{isensee2021nnu}          & 77.98$\pm$0.42 & 85.91$\pm$0.57 & 9.00$\pm$0.63  & 1.14$\pm$0.05 \\
MedNeXt~\cite{roy2023mednext}          & 77.20$\pm$0.14 & 84.69$\pm$0.09 & 9.56$\pm$0.35  & 1.19$\pm$0.03 \\
\hline
UNETR~\cite{hatamizadeh2022unetr}      & 76.09$\pm$0.41 & 82.51$\pm$0.38 & 9.55$\pm$0.76  & 1.22$\pm$0.07 \\
Swin UNETR~\cite{hatamizadeh2021swin}  & 77.71$\pm$0.46 & 85.36$\pm$0.48 & 9.26$\pm$0.52  & 1.16$\pm$0.04 \\
SwinUNETR-V2~\cite{he2023swinunetr}    & 78.03$\pm$0.48 & 85.61$\pm$0.67 & 9.13$\pm$0.73  & 1.12$\pm$0.06 \\
nnFormer~\cite{zhou2021nnformer}       & 74.90$\pm$0.62 & 82.74$\pm$0.63 & 10.93$\pm$0.67 & 1.37$\pm$0.08 \\
\hline
NA-UNETR (no-dil.)                & 78.74$\pm$0.29 & 86.13$\pm$0.36 & 9.05$\pm$0.44  & 1.14$\pm$0.04 \\
NA-UNETR (ours)                        & \textbf{79.49$\pm$0.25} & \textbf{86.88$\pm$0.32} & \textbf{8.89$\pm$0.30} & \textbf{1.02$\pm$0.03} \\
\hline
\hline
\end{tabular}
\end{center}
\end{table}

\autoref{tab:main_results} and \autoref{tab:main_results2} summarize the performance of NA-UNETR~(ours) and NA-UNETR~(no-dil.), the no-dilation variant of the model, compared with state-of-the-art CNN and Transformer baselines. LAD-SEG serves as the primary evaluation under challenging non-contrast conditions, with ImageCAS included to demonstrate performance on a larger, high-contrast coronary dataset.

On LAD-SEG, CNN-based models achieved limited accuracy, with Dice scores between 35--43\%, HD95 values often above 40~mm, and ASD greater than 10~mm. Transformer-based methods performed better, with Swin UNETR reaching 44.78\% Dice and SwinUNETR-V2 achieving an ASD of 10.18~mm. NA-UNETR obtained the strongest overall results with 45.64\% Dice, 44.39\% clDice, 38.16~mm HD95, and 10.01~mm ASD, representing improvements in both volumetric overlap and centerline topology. Relative to its no-dilation variant, NA-UNETR achieved higher Dice and clDice and reduced HD95, indicating that dilated NA blocks help capture longer vessel trajectories. These gains correspond to approximately a 3\% Dice improvement over nnU-Net and a 3~mm reduction in HD95 relative to Swin UNETR. Mannâ€“Whitney U testing indicated that the differences were not statistically significant ($p > 0.05$), which is expected given the small sample size (n = 20) and the substantial inter-patient variability in LAD-SEG.

On ImageCAS, CNN-based models performed strongly, with UNet++ achieving 78.57\% Dice, 84.71\% clDice, 9.46~mm HD95, and 1.15~mm ASD. Transformer-based approaches such as SwinUNETR-V2 reached 78.03\% Dice, 85.61\% clDice, and 1.12~mm ASD. NA-UNETR surpassed both groups, achieving 79.49\% Dice, 86.88\% clDice, 8.89~mm HD95, and 1.02~mm ASD. Compared with its no-dilation variant, NA-UNETR showed modest but consistent gains across Dice, clDice, HD95, and ASD, indicating improved robustness even under high-contrast imaging. These gains correspond to a 1.2\% Dice improvement over UNet++, a 4\% reduction in HD95 compared with Swin UNETR, and roughly a 9\% reduction in ASD relative to the next best model, together with the highest centerline accuracy. Mannâ€“Whitney U testing confirmed that NA-UNETRâ€™s improvements on ImageCAS were statistically significant ($p < 0.05$).

\subsection{Ablation Studies}

\begin{table}[htbp]
\begin{center}
\captionv{15}{tab:ablation1}{Ablation studies: various architecture settings.
\label{tab:ablation1}
\vspace*{2ex}
}
\scriptsize
    \begin{tabular}{|ccc|ccc|}
        \hline
        NAT Blocks&Residual&Kernel&\multirow{2}{*}{\textbf{DSC (\%)}$\uparrow$} &\multirow{2}{*}{\textbf{HD95 (mm)}$\downarrow$} & \multirow{2}{*}{\textbf{ASD (mm)}$\downarrow$} \\
        Per Stage&Block&Size &&&\\
        \hline \hline
        (3, 4, 6, 18, 5)  & $\checkmark$ & variable  &45.64 & 38.16 &10.01\\
        (2, 3, 4, 18, 5)  & $\checkmark$ & variable  & 44.61& 41.63&11.10\\
        (2, 2, 2, 2, 2) & $\checkmark$ & variable   & 43.26&40.60 &10.93\\
        (3, 4, 6, 18, 5) &  $\checkmark$& static   & 43.17& 40.20&10.84\\
        (3, 4, 6, 18, 5) &  $\times$& variable  & 43.01& 42.08&11.43\\
        (3, 4, 6, 18, 5) &  $\times$& static  &42.28 & 40.67&10.80\\
        \hline
    \end{tabular}
 
\end{center}
\end{table}

\noindent\textbf{Architecture Ablations.} Ablation results in \autoref{tab:ablation1} show the effect of several key design choices in NA-UNETR. The default configuration uses a non-uniform NAT depth of $(3,4,6,18,5)$ with residual blocks preceding each NAT block and variable kernel sizes across stages. Reducing the NAT depth to $(2,3,4,18,5)$, as proposed in the original Neighborhood Attention work \cite{hassani2023neighborhood}, lowered DSC by approximately 1.0\%. A uniform allocation of $(2,2,2,2,2)$, similar to Swin UNETR \cite{hatamizadeh2021swin}, produced an even larger reduction of about 2.4\%. Removing residual blocks further decreased DSC from 45.64\% to 43.01\% and increased ASD, confirming their role in stabilizing local feature extraction. Using a fixed $3\times3\times3$ kernel also reduced DSC by roughly 2.5\%, indicating that variable kernels better accommodate stage-specific receptive field requirements. 

\begin{table}[htbp]
\begin{center}
\captionv{16}{tab:lora_rank}{Ablation study on LoRA rank $r$.
\label{tab:lora_rank}
\vspace*{2ex}
}
\scriptsize
    \begin{tabular}{|c|ccc|}
        \hline
        \multirow{2}{*}{\textbf{Rank $r$}} & 
        \multirow{2}{*}{\textbf{DSC (\%)}$\uparrow$} & 
        \multirow{2}{*}{\textbf{HD95 (mm)}$\downarrow$} & 
        \multirow{2}{*}{\textbf{ASD (mm)}$\downarrow$} \\
        &&&\\
        \hline \hline
        2  & 44.32 & 41.37 & 11.02 \\
        4  & 45.18 & 40.32 & 10.41 \\
        8  & \textbf{45.64} & \textbf{38.16} & \textbf{10.01} \\
        16 & 45.41 & 40.08 & 10.22 \\
        \hline
    \end{tabular}
\end{center}
\end{table}

\noindent\textbf{LoRA Rank Sensitivity.}
We evaluate LoRA rank $r \in \{2,4,8,16\}$ while keeping all other settings fixed. As shown in \autoref{tab:lora_rank}, performance improves from $r=2$ to $r=8$, with higher DSC and lower HD95 and ASD. The results suggest that increasing the rank improves the model's adaptation capacity for capturing fine-grained vessel structures, whereas too small a rank limits expressiveness. Increasing the rank beyond 8 does not yield further improvement and slightly degrades performance, indicating diminishing returns. Therefore, $r=8$ is adopted in all experiments.


\begin{table}[htbp]
\begin{center}
\captionv{15}{tab:ablation1}{Ablation on training strategy components.}
\label{tab:ablation2}
\vspace*{2ex}

\centering
\small
\setlength{\tabcolsep}{10pt}
\resizebox{0.70\columnwidth}{!}{
\begin{tabular}{|l|ccc|}
\hline
Variant & DSC$\uparrow$ & HD95 (mm)$\downarrow$ & ASD (mm)$\downarrow$ \\
\hline
NA-UNETR (default)  & \textbf{45.64} & \textbf{38.16} & \textbf{10.01} \\
\quad Train on LAD-SEG only & 36.39 & 40.72 & 10.90 \\
\quad Static loss weights & 44.08 & 42.19 & 11.35 \\
\quad No Focal Loss & 44.01 & 41.70 & 10.88 \\
\quad Segmentation loss only & 43.47 & 42.49 & 11.93 \\
\quad Standard preprocessing only & 43.12 & 40.91 & 11.44 \\
\quad No postprocessing & 44.41 & 40.74 & 10.47 \\
\hline
nnU-Net (standard preprocessing) & 42.54 & 39.68 & 10.37 \\
\quad Standard Preprocessing Only & 39.98 & 41.22 & 11.50 \\
\hline
\end{tabular}}
\end{center}
\end{table}

\noindent\textbf{Training Strategy Ablations.} Ablation results in Table~\ref{tab:ablation2} show that the two stage approach of pretraining on ImageCAS followed by fine tuning on LAD-SEG produced the strongest performance for NA-UNETR. Training on LAD-SEG alone reduced NA-UNETR DSC (45.64\% $\to$ 36.39\%), showing that pretraining provides a more stable initialization for thin vessel learning. Using fixed loss weights instead of dynamic weighting decreased DSC and increased boundary errors, and removing the boundary loss led to further degradation, indicating that both components contribute to balancing overlap and contour accuracy. Removing the focal term (Dice + Hausdorff) reduced DSC, and removing postprocessing (largest component, small object removal, hole filling) caused a modest drop, confirming these components provide incremental but non-dominant gains. Standard preprocessing reduced DSC and worsened boundary metrics for both NA-UNETR (45.64\% $\to$ 43.12\%) and nnU-Net (42.54\% $\to$ 39.98\%), with similar reductions of 2.52 and 2.56 percentage points respectively, indicating that the tailored pipeline provides consistent incremental benefits across architectures. Here, ``standard preprocessing'' denotes intensity clipping and linear scaling only.

    
 
\subsection{Computational Cost Analysis}

\begin{table}[htbp]
\begin{center}
\captionv{15}{tab:ablation1}{Computational cost analysis, NA-UNETR vs Benchmark Models.}
\label{tab:computational}
\vspace*{2ex}

\scriptsize
\begin{tabular}{|c|ccccc|}
\hline
\multirow{2}{*}{Method} & Training & Trainable & FLOPS & Inference & Peak\\
 & (s/epoch) & Params. (M) & (B) & (S) & VRAM (GB)\\
\hline
nnU-Net~\cite{isensee2021nnu} & 17.25 & 8.05 & 399.59 & 0.50 & 2.78\\
MedNeXt~\cite{roy2023mednext} & 25.98 & 3.91 & 121.17 & 4.68 & 4.91\\
UNETR~\cite{hatamizadeh2022unetr} & 18.13 & 20.13 & 480.87 & 1.17 & 3.91\\
Swin UNETR~\cite{hatamizadeh2021swin} & 22.21 & 19.71 & 300.00 & 2.13 & 5.49\\
NA-UNETR & 18.77 & 19.60 & 314.10 & 1.33 & 4.17\\
\hline
\end{tabular}

\end{center}
\end{table}

\autoref{tab:computational} summarizes the computational requirements of NA-UNETR relative to CNN- and Transformer-based baselines. CNN models such as nnU-Net and MedNeXt generally involve fewer parameters or lower FLOPs, whereas Transformer architectures incur higher complexity. NA-UNETR has 19.6M trainable parameters and 314.1B FLOPs, comparable to Swin UNETR (19.7M, 300.0B) and notably lower than UNETR (20.1M, 480.9B). In addition, NA-UNETR achieves competitive inference latency (1.33 s per volume) with moderate peak GPU memory usage (4.17 GB), remaining more memory-efficient than Swin UNETR while maintaining strong segmentation accuracy. Training time per epoch is also comparable across models, with NA-UNETR (18.77 s) remaining close to nnU-Net (17.25 s) and UNETR (18.13 s), indicating no additional training overhead.

\section{Discussion}

Accurately delineating the LAD on non-contrast CT is difficult because the vessel often blends with adjacent structures, reducing the visibility of its boundaries. Prior studies have shown that manual LAD annotations on non-contrast CT exhibit substantial inter-observer variability, with reported Dice values ranging from 0.10 to 0.53 \cite{kaderka2019geometric, morris2019cardiac, zhou2017cardiac, van2019development, duane2017cardiac}, and automated methods typically achieving 0.09--0.30, often requiring manual correction \cite{kaderka2019geometric, morris2019cardiac, zhou2017cardiac, van2019development}. In this setting, NA-UNETR achieved a mean DSC of 45.6\% on LAD-SEG, which compares favorably to prior reports and to the baselines evaluated in this study. The tailored preprocessing and training pipeline improved performance in the tested control settings. Under this common pipeline, NA-UNETR achieved the highest overall accuracy, consistent with a benefit from its integration of local and global context through Neighborhood Attention.

Pretraining on CTA provides a clear visualization of the coronary tree and allows the encoder to learn stable vessel features that form a useful foundation for subsequent fine-tuning. Although CTA differs from non-contrast CT, the decoder and the LoRA-adapted MLP layers help mitigate the modality shift and enable the model to transfer structural information while adapting to the target domain. Transformer-based models benefited from fine-tuning the MLP encoder layers together with the decoder, enabling better adaptation to non-contrast CT and yielding higher DSC than the CNN-based baselines \autoref{tab:main_results}, while clDice was less influenced since topological continuity can be maintained without additional encoder fine-tuning. Future work will incorporate explicit domain-adaptation or modality-alignment strategies to further reduce the CTA--non-contrast CT domain gap.

NA-UNETR showed favorable generalization by outperforming baselines on our LAD-SEG dataset and achieving competitive performance on the public artery dataset when trained directly on that benchmark. Across both datasets, NA-UNETR achieved stronger boundary-based metrics (HD95, ASD) in addition to higher Dice-based performance (DSC, clDice). Although boundary accuracy is particularly important for LAD segmentation due to the vessel's small caliber and thin, elongated course, prior LAD-focused studies \cite{kaderka2019geometric, morris2019cardiac, zhou2017cardiac, van2019development} did not report boundary-based metrics, limiting their assessment of contour quality and topological continuity.

Despite these gains, boundary-based metrics remain challenging on LAD-SEG. The thin vessel caliber and weak boundaries often produce small, fragmented predictions that inflate HD95 and ASD even when the centerline is captured accurately, a pattern consistent across all models reflecting the intrinsic visibility limits of the LAD on a small non-contrast CT dataset. NA-UNETR achieved substantially better HD95 and ASD on ImageCAS, where the CCTA images provide clearer vessel definition, highlighting the strong dependence of boundary accuracy on imaging modality. On LAD-SEG, given that reported inter-observer Dice scores range from 0.10 to 0.53~\cite{duane2017cardiac}, the achieved DSC lies within expert variability, suggesting potential utility in reducing manual editing effort and improving consistency for small, elongated vascular structures. Prospective workflow studies evaluating time savings, edit burden, and user acceptance will be required to further establish its clinical impact.

Visual inspection in \autoref{fig:visual} and \autoref{fig:multiviews} showed that most errors arose when the vessel narrowed to only a few voxels or when local contrast was very weak, leading to small gaps or minor boundary offsets in the predictions. Although NA-UNETR shows improvements over baseline methods across multiple metrics, LAD segmentation on non-contrast CT remains fundamentally limited by the visibility of the anatomy, and model performance is ultimately bounded by the information available in the underlying images. These results represent a step forward, but the approach is not yet suitable for clinical deployment. Further refinement, larger annotated datasets, and validation across multiple institutions will be essential to establish the reliability and robustness required for routine clinical use.

\section{Conclusion}
The LAD artery's small caliber, poor soft-tissue contrast, and limited annotated data make it one of the most challenging targets for automated segmentation in non-contrast CT. This work addresses these difficulties through neighborhood attention for geometrically coherent local-global context modeling and uncertainty-guided loss balancing for precise boundary delineation. Generalizability to the small-data regime is further strengthened through pretraining on a broader coronary dataset followed by parameter-efficient fine-tuning. Together, these contributions advance the feasibility of automated LAD delineation and lay groundwork for its integration into cardiac dose assessment workflows in thoracic radiotherapy planning.

\section*{Acknowledgments}
This work was supported in part by a grant from Varian Medical Systems (Siemens Healthineers).

\section*{Conflict of Interest}
The authors declare that they have no conflicts of interest.

\section*{Data Availability Statement}
Code has been made available at \url{https://github.com/rafiibnsultan/NA_UNETR}.


\section*{References}
\addcontentsline{toc}{section}{\numberline{}References}
\vspace*{-20mm}











\end{document}